\documentclass[fleqn,10pt]{wlscirep}
\usepackage[utf8]{inputenc}
\usepackage[T1]{fontenc}
\usepackage{amsmath}

\newcommand{\h}[1]{\underline{\textbf{#1}}}

\title{Emerging Opportunities of Using Large Language Models for Translation Between Drug Molecules and Indications}

\author[1,+]{David Oniani}
\author[1,+]{Jordan Hilsman}
\author[2,3]{Chengxi Zang}
\author[4]{Junmei Wang}
\author[4]{Lianjin Cai}
\author[5]{Jan Zawala}
\author[1,6,7,8,*]{Yanshan Wang}

\affil[1]{University of Pittsburgh, Department of Health Information Management, Pittsburgh, PA, USA}
\affil[2]{Weill Cornell Medicine, Department of Population Health Sciences, New York, NY, USA}
\affil[3]{Weill Cornell Medicine, Institute of Artificial Intelligence for Digital Health, New York, NY, USA}
\affil[4]{University of Pittsburgh, Department of Pharmaceutical Sciences, Pittsburgh, PA, USA}
\affil[5]{Jerzy Haber Institute of Catalysis and Surface Chemistry, Polish Academy of Sciences, Kraków, Poland}
\affil[6]{Department of Biomedical Informatics, University of Pittsburgh, Pittsburgh, PA, USA}
\affil[7]{Intelligent Systems Program, University of Pittsburgh, Pittsburgh, PA, USA}
\affil[8]{Clinical and Translational Science Institute, University of Pittsburgh, Pittsburgh, PA, USA}

\affil[+]{these authors contributed equally to this work}
\affil[*]{yanshan.wang@pitt.edu}

\keywords{Computer Science, Artificial Intelligence, Large Language Models, Drug Discovery}

\begin{abstract}
    A drug molecule is a substance that changes the organism's mental or physical state. Every
    approved drug has an indication, which refers to the therapeutic use of that drug for treating a particular
    medical condition. While the Large Language Model (LLM), a generative Artificial Intelligence (AI) technique, has recently demonstrated effectiveness in translating between molecules and their textual descriptions, there remains a gap in research regarding their application in facilitating the translation between drug molecules and indications, or vice versa, which could greatly benefit the drug discovery process. The capability of generating a drug from a given indication
    would allow for the discovery of drugs targeting specific diseases or targets and ultimately provide
    patients with better treatments. In this paper, we first propose a new task, which is the translation
    between drug molecules and corresponding indications, and then test existing LLMs on this new task. Specifically, we consider nine variations of the T5 LLM and
    evaluate them on two public datasets obtained from ChEMBL and DrugBank. Our experiments show the early
    results of using LLMs for this task and provide a perspective on the state-of-the-art. We also
    emphasize the current limitations and discuss future work that has the potential to improve the
    performance on this task. The creation of molecules from indications, or vice versa, will allow
    for more efficient targeting of diseases and significantly reduce the cost of drug discovery, with the potential to
    revolutionize the field of drug discovery in the era of generative AI.
\end{abstract}

\begin{document}

\flushbottom
\maketitle
\thispagestyle{empty}


\section*{Introduction}

Drug discovery is a costly process~\cite{wouters2020} that identifies chemical entities with the
potential to become therapeutic agents~\cite{decker2007}. Due to its clear benefits and significance
to health, drug discovery has become an active area of research, with researchers attempting to
automate and streamline drug discovery~\cite{schneider2018,sadybekov2023}. Approved drugs have indications, which refer to the use of that drug for treating a particular
disease~\cite[Chapter~5]{mehta2008}. The creation of molecules from indications, or vice versa, will allow
    for more efficient targeting of diseases and significantly reduce the cost of drug discovery, with the potential to
    revolutionize the field of drug discovery.

Large Language Models (LLMs) have become one of the major directions of generative Artificial Intelligence (AI)
research, with highly performant models like GPT-3~\cite{brown2020}, GPT-4~\cite{openai2023},
LLaMA~\cite{touvron2023}, and Mixtral~\cite{jiang2024} developed in the recent years and services
like ChatGPT reaching over 100 million users~\cite{chatgptverge,chatgptreuters}. LLMs utilize deep
learning methods to perform various Natural Language Processing (NLP) tasks, such as text
generation~\cite{chung2023,lee2022} and neural machine translation~\cite{moslem2023,mu2023}. The
capabilities of LLMs are due in part to their training on large-scale textual data, making the
models familiar with a wide array of topics. LLMs have also demonstrated promising performance in a
variety of tasks across different scientific
fields~\cite{singhal2023,yu2023,gomez-rodriguez2023,boiko2023}. Since LLMs work with textual data,
the first step is usually finding a way to express a problem in terms of text or language.

An image or a diagram is a typical way to present a molecule, but methods for obtaining textual
representations of molecules do exist. One such method is the
Simplified~Molecular-Input~Line-Entry~System~(SMILES)~\cite{weininger1988}, which is usually considered as a
language for describing molecules. As SMILES strings represent drugs in textual form, we can assess
the viability of LLMs in translating between drug molecules and their indications. In this paper, we consider
two tasks: drug-to-indication and indication-to-drug, where we seek to generate indications from the
SMILES strings of existing drugs, and SMILES strings from indications, respectively. Translation
between drugs and the corresponding indication will allow for finding a cure for diseases that have no
current treatment, and give clinicians more avenues for patient care. We also release the codebase
for the study\footnote{\url{https://github.com/PittNAIL/drug-to-indication}}.

Research efforts have attempted de-novo drug discovery through the use of AI and, more recently, forms
of generative AI~\cite{debleena2021}. There are numerous existing efforts for molecular design and
drug discovery using AI, such as GPT-based models using scaffold SMILES strings accompanied with
desired properties of the output molecule~\cite{bagal2022}. Others have used T5 architecture for
various tasks, such as reaction prediction~\cite{lu2022} and converting between molecular captions
and SMILES strings~\cite{edwards2022}. Additional work in the field is centered around the
generation of new molecules from gene expression signatures using generative adversarial
networks~\cite{mendez-lucio2020}, training recurrent neural networks on known compounds and their
SMILES strings, then fine-tuning for specific agonists of certain receptors~\cite{merk2018}, or
using graph neural networks to predict drugs and their corresponding indications from
SMILES~\cite{smilegnn}. As such, there is an established promise in using AI for drug discovery and
molecular design. Efforts to make data more friendly for AI generation of drugs include the
development of the Self-Referencing~Embedded~Strings~(SELFIES)~\cite{krenn2020}, which can represent
every valid molecule. The reasoning is that such a format will allow generative AI to construct valid
molecules while maintaining crucial structural information in the string. The collection of these
efforts sets the stage for our attempt at generating drug indications from molecules.

With advancements in medicinal chemistry leading to an increasing number of drugs designed for
complex processes, it becomes crucial to comprehend the distinctive characteristics and subtle
nuances of each drug. This has led to the development of molecular fingerprints, such as the Morgan
fingerprint~\cite{morgan1965} and the MAP4 fingerprint~\cite{capecchi2020}, which use unique
algorithms to vectorize the characteristics of a molecule. Computation of fingerprint
representations is rapid, and they maintain much of the features of a molecule~\cite{wigh2022}.
Molecular fingerprinting methods commonly receive input in the form of SMILES strings, which serve
as a linear notation for representing molecules in their structural forms, taking into account the
different atoms present, the bonds between atoms, as well as other key characteristics, such as
branches, cyclic structures, and aromaticity~\cite{weininger1988}. Since SMILES is a universal
method of communicating the structure of different molecules, it is appropriate to use SMILES
strings for generating fingerprints. Mol2vec~\cite{jaeger2018} feeds Morgan fingerprints to the
Word2vec~\cite{mikolov2013} algorithm by converting molecules into their textual representations.
BERT~\cite{bert}-based models have also been used for obtaining molecular representations, including
models like MolBERT~\cite{molbert} and ChemBERTa~\cite{chemberta}, which are pretrained BERT
instances that take SMILES strings as input and perform downstream tasks on molecular representation
and molecular property prediction, respectively. Other efforts in using AI for molecular
representations include generating novel molecular graphs through the use of reinforcement learning,
decomposition, and reassembly~\cite{yamada2023} and the prediction of 3D representations of small
molecules based on their 2D graphical counterparts~\cite{ganea2021}.

In this paper, we evaluate the capabilities of MolT5, a T5-based model, in translating between drugs
and their indications through the two tasks, drug-to-indication and indication-to-drug, using drug
data from DrugBank and ChEMBL. The drug-to-indication task utilizes SMILES strings for existing
drugs as input, with the matching indications of the drug as the target output. The
indication-to-drug task takes the set of indications for a drug as input and seeks to generate the
corresponding SMILES string for a drug that treats the listed conditions.

We employ all available MolT5 model sizes for our experiments and evaluate them separately across
the two datasets. Additionally, we perform the experiments under three different configurations:

\begin{enumerate}
    \item Evaluation of the baseline models on the entire available dataset
    \item Evaluation of the baseline models on 20\% of the dataset
    \item Fine-tuning the models on 80\% of the dataset followed by evaluation on the 20\% subset
\end{enumerate}

Larger MolT5 models outperformed the smaller ones across all configurations and tasks. It should
also be noted that fine-tuning MolT5 models has a negative impact on the performance.

Following these preliminary experiments, we train the smallest available MolT5 model from scratch
using a custom tokenizer. This custom model performed better on DrugBank data than on ChEMBL data on
the drug-to-indication task, perhaps due to a stronger signal between the drug indications and
SMILES strings in their dataset, owing to the level of detail in their indication descriptions.
Fine-tuning the custom model on 80\% of either dataset did not degrade model performance for either
task, and some metrics saw improvement due to fine-tuning. Overall, fine-tuning for the
indication-to-drug task did not consistently improve the performance, which holds for both ChEMBL
and DrugBank datasets.

While the performance of the custom tokenizer approach is still poor, there is promise in using a
larger model and having access to more data. If we have a wealth of high-quality data to train
models on translation between drugs and their indications, it may be possible to improve performance
and facilitate novel drug discovery with LLMs.


\section*{Results}

\subsection*{Evaluation of MolT5 Models}

We performed initial experiments using MolT5 models from
HuggingFace\footnote{\url{https://huggingface.co/laituan245/molt5-small/tree/main}}\footnote{\url{https://huggingface.co/laituan245/molt5-base/tree/main}}\footnote{\url{https://huggingface.co/laituan245/molt5-large/tree/main}}.
MolT5 offers three model sizes and fine-tuned models of each size, which support each task of our
experiments. For experiments generating SMILES strings from drug indications (drug-to-indication),
we used the fine-tuned models MolT5-smiles-to-caption, and for generating SMILES strings from drug
indications (indication-to-drug), we used the models MolT5-caption-to-smiles. For each of our
Tables, we use the following flags: FT (denotes experiments where we fine-tuned the models on 80\%
of the dataset and evaluated on the remaining 20\% test subset), SUB (denotes experiments where the
models are evaluated solely on the 20\% test subset), and FULL (for experiments evaluating the
models on the entirety of each dataset.

For evaluating drug-to-indication, we employ the natural language generation metrics
BLEU~\cite{papineni20002}, ROGUE~\cite{lin2004}, and METEOR~\cite{banerjee2005}, as well as the
Text2Mol~\cite{edwards2021} metric, which generates similarities of SMILES-Indication pairs. As for
evaluation of indication-to-drug, we measure exact SMILES string matches, Levenshtein
distance~\cite{miller2009}, SMILES BLEU scores, the Text2Mol similarity metric, as well as three
different molecular fingerprint metrics: MACCS, RDK, and Morgan FTS, where FTS stands for
fingerprint Tanimoto similarity~\cite{tanimoto1958}, as well as the proportion of returned SMILES
strings that are valid molecules. The final metric for evaluating SMILES generation is FCD, or
Fréchet ChemNet Distance, which measures the distance between two distributions of molecules from
their SMILES strings~\cite{preuer2018}.

\begin{table}[hbt!]
  \centering
  \begin{tabular}{|l|c|c|c|c|c|c|c|}
    \hline
    Model      & BLEU-2     & BLEU-4     & ROUGE-1    & ROUGE-2    & ROUGE-L    & METEOR     & Text2Mol\\
    \hline
    FT-Small   &    0.0145  &    0.0020  &    0.0847  &    0.0046  &    0.0699  &    0.0887  & \h{0.3599}\\
    SUB-Small  & \h{0.0225} & \h{0.0056} & \h{0.0966} & \h{0.0064} & \h{0.0801} & \h{0.0991} &    0.3179\\
    FULL-Small &    0.0210  &    0.0036  &    0.0957  &    0.0059  &    0.0793  &    0.0964  &    0.3251\\
    \hline
    FT-Base    &    0.0156  &    0.0020  &    0.0874  &    0.0048  &    0.0728  &    0.0922  &    0.2444\\
    SUB-Base   & \h{0.0233} & \h{0.0055} & \h{0.0980} & \h{0.0069} & \h{0.0828} & \h{0.0987} &    0.2740\\
    FULL-Base  &    0.0218  &    0.0035  &    0.0970  &    0.0058  &    0.0806  &    0.0974  & \h{0.3267}\\
    \hline
    FT-Large   &    0.0247  &    0.0065  &    0.0998  &    0.0084  & 0.0811     &    0.1097  & \h{0.4815}\\
    SUB-Large  & \h{0.0310} & \h{0.0103} & \h{0.1045} & \h{0.0132} & \h{0.0845} & \h{0.1241} &    0.4734\\
    FULL-Large &    0.0294  &    0.0093  &    0.1019  &    0.0114  & 0.0817     &    0.1172  &    0.4788\\
    \hline
  \end{tabular}
  \caption{\label{tab:drugbank_drug-to-indication} DrugBank drug-to-indication Results.}
\end{table}

\begin{table}[hbt!]
  \centering
  \begin{tabular}{|l|c|c|c|c|c|c|c|}
    \hline
    Model      & BLEU-2     & BLEU-4     & ROUGE-1    & ROUGE-2    & ROUGE-L    & METEOR     & Text2Mol\\
    \hline
    FT-Small   &    0.0003  &    0.0000  &    0.0007  &    0.0000  &    0.0007  &    0.0033  &    0.1367\\
    SUB-Small  &    0.0020  &    0.0000  &    0.0650  &    0.0000  &    0.0617  &    0.0619  & \h{0.3527}\\
    FULL-Small & \h{0.0023} &    0.0000  & \h{0.0672} &    0.0000  & \h{0.0635} & \h{0.0628} &    0.3247\\
    \hline
    FT-Base    &    0.0009  &    0.0000  &    0.0452  &    0.0000  &    0.0426  &    0.0371  &     0.0796\\
    SUB-Base   &    0.0019  &    0.0000  & \h{0.0667} &    0.0000  & \h{0.0629} &    0.0622  &  \h{0.3335}\\
    FULL-Base  & \h{0.0023} & \h{0.0013} &    0.0592  & \h{0.0009} &    0.0562  & \h{0.0709} &    0.3222\\
    \hline
    FT-Large   & \h{0.0100} & \h{0.0038} &    0.0503  & \h{0.0340} &    0.0466  & \h{0.0770} &    0.1443\\
    SUB-Large  &    0.0056  &    0.0010  &    0.0584  &    0.0008  &    0.0552  &    0.0701  & \h{0.4841}\\
    FULL-Large &    0.0062  &    0.0012  & \h{0.0592} &    0.0009  & \h{0.0562} &    0.0709  &    0.4673\\
    \hline
  \end{tabular}
  \caption{\label{tab:chembl_drug-to-indication} ChEMBL drug-to-indication Results.}
\end{table}

Tables \ref{tab:drugbank_drug-to-indication} and \ref{tab:chembl_drug-to-indication} show the
results of MolT5 experiments on DrugBank and ChEMBL data for drug-to-indication, respectively.
Larger models tended to perform better across all metrics for each experiment. Across almost all
metrics for the drug-to-indication task, on both DrugBank and ChEMBL datasets, the model performed
best on the 20\% subset data. At the same time, both the subset and full dataset evaluations yielded
better results than fine-tuning experiments. As MolT5 models are trained on molecular captions,
fine-tuning using indications could introduce noise and weaken the signal between input and target
text. The models performed better on DrugBank data than ChEMBL data, which may be due to the level
of detail provided by DrugBank for their drug indications.

\begin{table}[hbt!]
  \centering
  \begin{tabular}{|l|c|c|c|c|c|c|c|c|c|c|}
    \hline
    Model & BLEU\(\uparrow\) & Exact\(\uparrow\) & Levenshtein\(\downarrow\) & MACCS & RDK & Morgan & FCD & Text2Mol & Validity\\
    \hline
    FT-Small   &    0.0982  &    0.0000  &    167.9400  & \h{0.3336} & \h{0.2037} &    0.0877  & \h{42.2532}   &    0.0000  &    0.0169\\
    SUB-Small  &    0.1196  &    0.0000  &    123.1015  &    0.2841  &    0.1792  &    0.0901  &    21.4590    &    0.0674  & \h{0.2192}\\
    FULL-Small & \h{0.1329} &    0.0000  & \h{116.5606} &    0.2972  &    0.1816  & \h{0.0962} &    13.9813    & \h{0.0504} &    0.2116\\
    \hline
    FT-Base    &    0.0207  &    0.0000  &    452.3344  &    0.0159  &    0.1505  &    0.0000  & \h{44.2047}   &    0.0000  &    0.0091\\
    SUB-Base   &    0.1275  &    0.0000  & \h{118.1165} &    0.3122  &    0.1937  &    0.1122  &    22.3573    & \h{0.1381} & \h{0.2062}\\
    FULL-Base  & \h{0.1337} & \h{0.0003} &    120.7320  & \h{0.3350} & \h{0.2111} & \h{0.1191} &    15.1980    &    0.0701  &    0.1931\\
    \hline
    FT-Large   &    0.0284  &    0.0000  &    412.8220  &    0.1417  &    0.1061  &    0.0348  & \h{45.1149}   &   -0.1357  &    0.0117\\
    SUB-Large  &    0.0959  & \h{0.0233} & \h{181.4176} &    0.4104  &    0.2976  &    0.1732  &    12.5943    & \h{0.2588} & \h{0.5279}\\
    FULL-Large & \h{0.0999} &    0.0203  &    182.3688  & \h{0.4192} & \h{0.3090} & \h{0.1771} &    8.2989     &    0.2314  &    0.4999\\
    \hline
  \end{tabular}
  \caption{\label{tab:drugbank_indication-to-drug} DrugBank indication-to-drug Results.}
\end{table}

\begin{table}[hbt!]
  \centering
  \begin{tabular}{|l|c|c|c|c|c|c|c|c|c|c|}
    \hline
    Model & BLEU\(\uparrow\) & Exact\(\uparrow\) & Levenshtein\(\downarrow\) & MACCS & RDK & Morgan & FCD & Text2Mol & Validity\\
    \hline
    FT-Small   &    0.0080  & 0.0000 &    461.1794  &    0.0000  &    0.0000  &    0.0000  &    0.0000   &  \h{0.5462} &    0.0000\\
    SUB-Small  &    0.0996  & 0.0000 &    162.1305  &    0.2635  & \h{0.1467} & \h{0.0676} &    0.0000   &    -0.0377  & \h{0.2618}\\
    FULL-Small & \h{0.1021} & 0.0000 & \h{161.5838} & \h{0.2638} &    0.1446  &    0.0666  & \h{24.5937} &    -0.0183  &    0.2554\\
    \hline
    FT-Base    &    0.0245  & 0.0000 &    490.3377  &    0.0000  &    0.0000  &    0.0000  &    0.0000   & \h{0.5462}  &    0.0091\\
    SUB-Base   & \h{0.1737} & 0.0000 &    87.8989   & \h{0.3224} & \h{0.2008} & \h{0.0927} &    0.0000   &    0.0560   &    0.3605\\
    FULL-Base  &    0.1727  & 0.0000 & \h{86.9512}  &    0.3122  &    0.2004  &    0.0884  & \h{27.7150} &    0.0539   & \h{0.3723}\\
    \hline
    FT-Large   &    0.0077  &    0.0000  &    414.8030  &    0.0169  &    0.0000  &    0.0007  &    0.0000   &         0.1600   &    0.0703\\
    SUB-Large  &    0.0581  & \h{0.0008} &    355.9959  &    0.3837  &    0.2953  &    0.0781  &    0.0000   &         0.1206   &    0.3287\\
    FULL-Large & \h{0.0599} &    0.0002  & \h{349.3339} & \h{0.3978} & \h{0.3035} & \h{0.0810} & \h{24.7671} & \h{0.0733}  & \h{0.3516}\\
    \hline
  \end{tabular}
  \caption{\label{tab:chembl_indication-to-drug} ChEMBL indication-to-drug Results.}
\end{table}

Tables \ref{tab:drugbank_indication-to-drug} and \ref{tab:chembl_indication-to-drug} show the
results of MolT5 experiments on DrugBank and ChEMBL data for indication-to-drug, respectively. The
tables indicate that fine-tuning the models on the new data worsens performance, reflected in FT
experiments yielding worse results than SUB or FULL experiments. Also, larger models tend to perform
better across all metrics for each experiment.

In our drug-to-indication and indication-to-drug experiments, we see that fine-tuning the models
causes the models to perform worse across all metrics. Additionally, larger models perform better on
our tasks. However, in our custom tokenizer experiments, we pretrain MolT5-Small without the added
layers of Smiles-to-Caption and Caption-to-Smiles. By fine-tuning the custom pretrained model on
our data for drug-to-indication and indication-to-drug, we aim to see improved results.

\subsection*{Evaluation of Custom Tokenizer}

\begin{table}[hbt!]
  \centering
  \begin{tabular}{|l|c|c|c|c|c|c|c|}
    \hline
    Model         & BLEU-2 & BLEU-4 & ROUGE-1 & ROUGE-2 & ROUGE-L & METEOR & Text2Mol\\
    \hline
    FT-DrugBank   & \h{0.0006} & 0.0000 & \h{0.0013}  & 0.0000  & \h{0.0013}  & \h{0.0141} &    0.0657\\
    FT-ChEMBL     &    0.0000  & 0.0000 &    0.0011   & 0.0000  &    0.0011   &    0.0017  & \h{0.0931}\\
    \hline
    SUB-DrugBank  & \h{0.0008} & 0.0000 & \h{0.0014}  & 0.0000  & \h{0.0013}  & \h{0.0137} &    0.0633\\
    SUB-ChEMBL    &    0.0000  & 0.0000 &    0.0012   & 0.0000  &    0.0012   &    0.0012  & \h{0.0733}\\
    \hline
    FULL-DrugBank & \h{0.0010} & 0.0000 &    0.0014   & 0.0000  &    0.0014   & \h{0.0133} &    0.0679\\
    FULL-ChEMBL   &    0.0000  & 0.0000 & \h{0.0016}  & 0.0000  & \h{0.0016}  &    0.0014  & \h{0.0709}\\
    \hline
  \end{tabular}
  \caption{\label{tab:custom_drug-to-indication}Results for MolT5 augmented with custom tokenizer, drug-to-indication.}
\end{table}

\begin{table}[hbt!]
  \centering
  \begin{tabular}{|l|c|c|c|c|c|c|c|c|c|c|}
    \hline
    Model & BLEU\(\uparrow\) & Exact\(\uparrow\) & Levenshtein\(\downarrow\) & MACCS & RDK  & Morgan & FCD & Text2Mol & Validity\\
    \hline
    FT-DrugBank   & \h{0.0101} & 0.0000 & \h{228.0395} &    0.0048  &    0.0007  &    0.0017  & 0.0000 & -0.2507 & \h{0.0815}\\
    FT-ChEMBL     &    0.0022  & 0.0000 &    853.0084  & \h{0.0280} & \h{0.0292} & \h{0.0052} & 0.0000 & 0.0000  &    0.0049\\
    \hline
    SUB-DrugBank  & \h{0.0124} & 0.0000 & \h{176.4470} & \h{0.0024} &   0.0004  & \h{0.0022} & 0.0000 &    0.0000   & \h{0.0058}\\
    SUB-ChEMBL    &    0.0090  & 0.0000 &    230.6411  &    0.0007 & \h{0.0186} &    0.0004  & 0.0000 & \h{0.2006}  &    0.0041\\
    \hline
    FULL-DrugBank & \h{0.0115} & 0.0000 & \h{185.8035} & \h{0.0034} & \h{0.0155} & \h{0.0018} & 0.0000 &    0.0928   &    0.0102\\
    FULL-ChEMBL   & 0.0087     & 0.0000 &    228.9100  &    0.0017  &    0.0075  &    0.0016  & 0.0000 & \h{0.1861}  & \h{0.0922}\\
    \hline
  \end{tabular}
  \caption{\label{tab:custom_indication-to-drug}Results for MolT5 augmented with custom tokenizer, indication-to-drug.}
\end{table}

Tables \ref{tab:custom_drug-to-indication} and \ref{tab:custom_indication-to-drug} show the
evaluation of MolT5 pretrained with the custom tokenizer on the drug-to-indication and
indication-to-drug tasks, respectively. For drug-to-indication, the model performed better on
the DrugBank dataset, reflected across all metrics. This performance difference may be due to a
stronger signal between drug indication and SMILES strings in their dataset, as their drug
indication text goes into great detail. Fine-tuning the model on 80\% of either of the datasets did
not worsen the performance for drug-to-indication as it did in the baseline results, and some
metrics showed improved results. The results for indication-to-drug are more mixed. The model does
not consistently perform better across either dataset and fine-tuning the model affects the
evaluation metrics inconsistently.


\section*{Discussion}

In this paper, we proposed a novel task of translating between drugs and indications, considering
both drug-to-indication and indication-to-drug subtasks. We focus on generating indications from the
SMILES strings of existing drugs and generating SMILES strings from sets of indications. Our
experiments are the first attempt at tackling this problem. After conducting experiments with
various model configurations and two datasets, we hypothesized potential issues that need further
work. We believe that properly addressing these issues could significantly improve the performance
of the proposed tasks.

The signal between SMILES strings and indications is poor. In the original MolT5 task (translation
between molecules and their textual descriptions), "similar" SMILES strings often had similar
textual descriptions. In the case of drug-to-indication and indication-to-drug tasks, similar SMILES
strings might have completely different textual descriptions as they are different drugs, and their
indications also differ. One could also make a similar observation about SMILES strings that are
different: drug indications may be similar. Having no direct relationships between drugs and
indications makes it hard to achieve high performance on proposed tasks. We hypothesize that having
an intermediate representation that drugs (or indications) map to may improve the performance. As an
example, mapping a SMILES string to its caption (MolT5 task) and then caption to indication may be a
potential future direction of research.

The signal between drugs and indications is not the only issue: the data is also scarce. Since we do
not consider random molecules and their textual descriptions but drugs and their indications, the
available data is limited by the number of drugs. In the case of both ChEMBL and DrugBank datasets,
the number of drug-indication pairs was under 10000, with the combined size also being under 10000.
Finding ways to enrich data may help establish a signal between SMILES strings and indications and
could be a potential future avenue for exploration.

Overall, the takeaway from our experiments is that the custom tokenizer approach has promise and
benefits from fine-tuning, but its performance has yet to meet our expectations. We also see in our
baseline experiments that larger models tend to perform better. By using a larger model and having
more data (or data that has a stronger signal between drug indications and SMILES strings), we may
be able to successfully translate between drug indications and molecules (i.e., SMILES strings) and
ultimately facilitate novel drug discovery.


\section*{Methods}

\begin{figure}[!htbp]
    \centering
    \includegraphics[width=\linewidth]{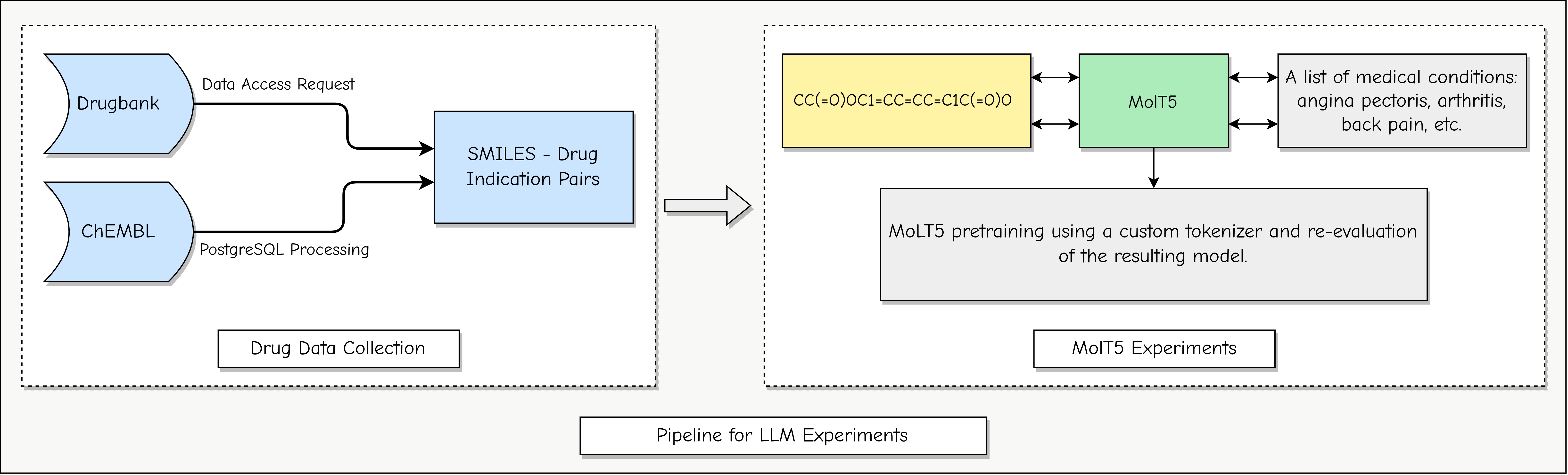}
    \caption{Overview of the methodology of the experiments: drug data is compiled from ChEMBL and
             DrugBank and utilized as input for MolT5. Our experiments involved two tasks:
             drug-to-indication and indication-to-drug. For drug-to-indication, SMILES strings of
             existing drugs were used as input, producing drug indications as output. Conversely,
             for drug-to-indication, drug indications of the same set of drugs were the input,
             resulting in SMILES strings as output. Additionally, we augmented MolT5 with a custom
             tokenizer in pretraining and evaluated the resulting model on the same tasks.}
    \label{fig:pipeline_overview}
\end{figure}

This section describes the dataset, analysis methods, ML models, and feature extraction techniques
used in this study. Figure ~\ref{fig:pipeline_overview} shows the flowchart of the process. We
adjust the workflow of existing models for generating molecular captions to instead generate
indications for drugs. By training LLMs on the connection between SMILES strings and drug
indications, we endeavor to one day be able to create novel drugs that treat medical conditions.

\subsection*{Data}

\begin{table}[hbt!]
  \centering
  \caption{Dataset Details}
  \label{tab:datasets}
  \begin{tabular}{|l|c|c|c}
  \hline
                                           & \textbf{DrugBank} & \textbf{ChEMBL}\\
    \hline
    Number of Drug-Indication Pairs        & 3004              & 6127\\
    Minimum Indication Length (Characters) & 19                & 34\\
    Minimum SMILES Length (Characters)     & 1                 & 1\\
    Average Indication Length (Characters) & 259               & 114\\
    Average SMILES Length (Characters)     & 59                & 67\\
    Maximum Indication Length (Characters) & 3517              & 524\\
    Maximum SMILES Length (Characters)     & 710               & 1486\\
    \hline
  \end{tabular}
\end{table}

Our data comes from two databases, DrugBank~\cite{wishart2006} and ChEMBL~\cite{davies2015}, which
we selected due to the different ways they represent drug indications. DrugBank gives in-depth
descriptions of how each drug treats patients, while ChEMBL provides a list of medical conditions
each drug treats. Table \ref{tab:datasets} outlines the size of each dataset, as well as the length
of the SMILES and Indication data. In the case of DrugBank, we had to request access to use the drug
indication and SMILES data. The ChEMBL data was available without request but required establishing
a database locally to query and parse the drug indication and SMILES strings into a workable format.
Finally, we prepared a pickle file for both databases to allow for metric calculation following the
steps presented in MolT5~\cite{edwards2022}.

\subsection*{Models}

We conducted initial experiments using the MolT5 model, based on the T5
architecture~\cite{edwards2022}. The T5 basis of the model gives it textual modality from
pretraining on the natural language text dataset Colossal Clean Crawled Corpus
(C4)~\cite{raffel2020}, and the pretraining on 100 million SMILES strings from the ZINC-15
dataset~\cite{zinc15} gives the model molecular modality.

In our experiments, we utilized fine-tuned versions of the available MolT5 models:
Smiles-to-Caption, fine-tuned for generating molecular captions from SMILES strings, and
caption-to-smiles, fine-tuned for generating SMILES strings from molecular captions. However, we
seek to evaluate the model's capacity to translate between drug indications and SMILES strings.
Thus, we use drug indications in the place of molecular captions, yielding our two tasks:
drug-to-indication and indication-to-drug.

The process of our experiments begins with evaluating the baseline MolT5 model for each task on the
entirety of the available data (3004 pairs for DrugBank, 6127 pairs for ChEMBL), on a 20\% subset of
the data (601 pairs for DrugBank, 1225 pairs for ChEMBL), and then fine-tuning the model on 80\%
(2403 pairs for DrugBank, 4902 pairs for ChEMBL) of the data and evaluating on that same 20\%
subset.

After compiling the results of the preliminary experiments, we decided to use a custom tokenizer
with the MolT5 model architecture. While the default tokenizer leverages the T5 pretraining, the
reasoning is that treating SMILES strings as a form of natural language and tokenizing it
accordingly into its component bonds and molecules could improve model understanding of SMILES
strings and thus improve performance.

\subsection*{MolT5 with Custom Tokenizer}

\begin{figure}[!htbp]
    \centering
    \includegraphics[width=\linewidth]{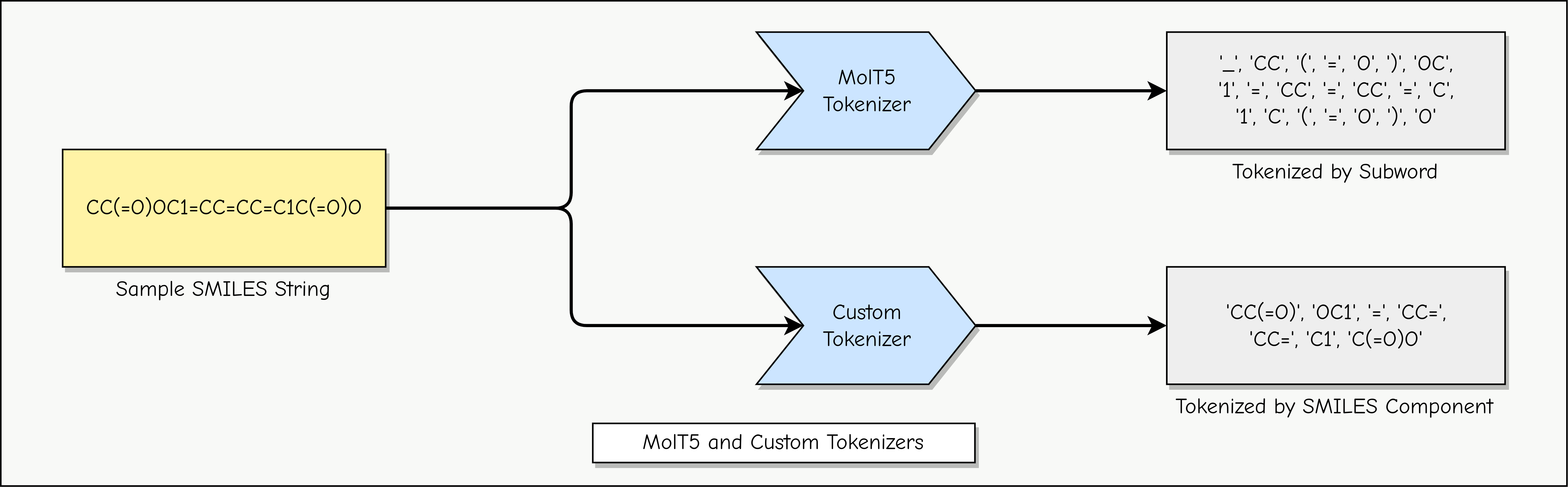}
    \caption{MolT5 and custom tokenizers: MolT5 tokenizer uses the default English language
             tokenization and splits the input text into subwords. The intuition is that SMILES
             strings are composed of characters typically found in English text, and pretraining on
             large-scale English corpora may be helpful. On the other hand, the custom tokenizer
             method utilizes the grammar of SMILES and decomposes the input into grammatically valid
             components.}
    \label{fig:tokenizer}
\end{figure}

The tokenizer for custom pretraining of MolT5 that we selected came from previous work on adapting
transformers for SMILES strings~\cite{adilov2021}. This tokenizer separates SMILES strings into
individual bonds and molecules. Figure~\ref{fig:tokenizer} illustrates the behavior of both MolT5
and custom tokenizers. Due to computational limits, we only performed custom pretraining of the
smallest available MolT5 model, with 77 million parameters. Our pretraining approach utilized the
model configuration of MolT5 and
JAX\footnote{\url{https://jax.readthedocs.io/en/latest/index.html}}/Flax\footnote{\url{https://github.com/google/flax}}
to execute the span-masked language model objective on the ZINC dataset~\cite{raffel2020}. Following
pretraining, we assessed model performance on both datasets. The experiments comprised three
conditions: fine-tuning on 80\% (2403 pairs for DrugBank, 4902 pairs for ChEMBL) of the data and
evaluating on the remaining 20\% (601 pairs for DrugBank, 1225 pairs for ChEMBL), evaluating on 20\%
of the data without fine-tuning, and evaluating on 100\% (3004 pairs for DrugBank, 6127 pairs for
ChEMBL) of the data.


\bibliography{main}


\section*{Author contributions}

D.O. led the study, designed the experiments, helped conduct the experiments, analyzed the results,
and wrote, reviewed, and revised the paper. J.H. contributed to the study design, conducted the
experiments, analyzed the results, and wrote, reviewed, and revised the paper. C.Z. shared
resources, contributed to discussions, and reviewed the paper. J.W. helped streamline the idea and
provided guidance from the perspective of chemistry. L.C. shared resources and contributed to
discussions. J.Z. contributed to discussions. Y.W. conceptualized the study and reviewed and revised
the paper.


\section*{Competing interests}

Y.W. consults for Pfizer Inc., and has ownership/equity interests in BonafideNLP, LLC. All other
authors declare no competing interests.


\section*{Additional information}

\textbf{Correspondence} and requests for materials should be addressed to Y.W.

\end{document}